\documentclass[smallabstract,smallcaptions]{dccpaper}

\usepackage{epsfig}
\usepackage{citesort}
\usepackage{amsmath}
\usepackage{amssymb}
\usepackage{color}
\usepackage{url}
\usepackage{multirow}
\usepackage[caption=false]{subfig}
\usepackage{graphicx}

\newlength{\figurewidth}
\newlength{\smallfigurewidth}

\makeatletter
\renewcommand\p@subfigure{\thefigure~}
\makeatother

\begin{document}

\title
{\large
\textbf{Exploring Compressed Image Representation as a Perceptual Proxy: A Study}
}

\author{%
Chen-Hsiu Huang and Ja-Ling Wu
\\[0.5em]
{\small\begin{minipage}{\linewidth}\begin{center}
\begin{tabular}{c}
Department of Computer Science and Information Engineering, \\
National Taiwan University \\
No. 1, Sec. 4, Roosevelt Rd., Taipei City 106, Taiwan\\
\url{{chenhsiu48,wjl}@cmlab.csie.ntu.edu.tw} 
\end{tabular}
\end{center}\end{minipage}}
}

\maketitle
\thispagestyle{empty}

\begin{abstract}
We propose an end-to-end learned image compression codec wherein the analysis transform is jointly trained with an object classification task. This study affirms that the compressed latent representation can predict human perceptual distance judgments with an accuracy comparable to a custom-tailored DNN-based quality metric. We further investigate various neural encoders and demonstrate the effectiveness of employing the analysis transform as a perceptual loss network for image tasks beyond quality judgments. Our experiments show that the off-the-shelf neural encoder proves proficient in perceptual modeling without needing an additional VGG network. We expect this research to serve as a valuable reference developing of a semantic-aware and coding-efficient neural encoder.
\end{abstract}

\Section{Introduction} \label{_Intro}

\begin{figure}[!ht]
\centering
\subfloat[]{{\includegraphics[height=5cm]{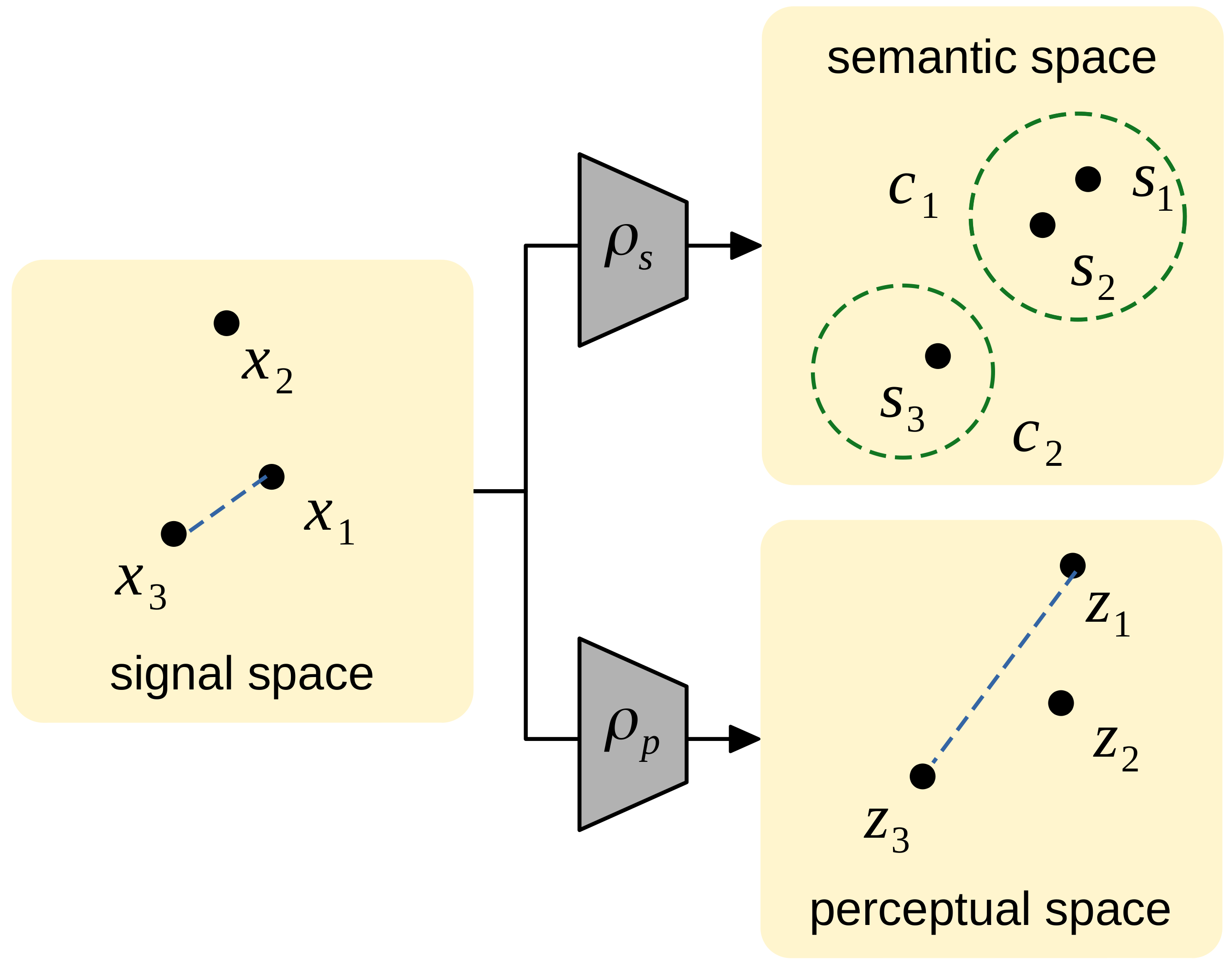} \label{fig:flow1}} }%
\hfill
\subfloat[]{{\includegraphics[height=5cm]{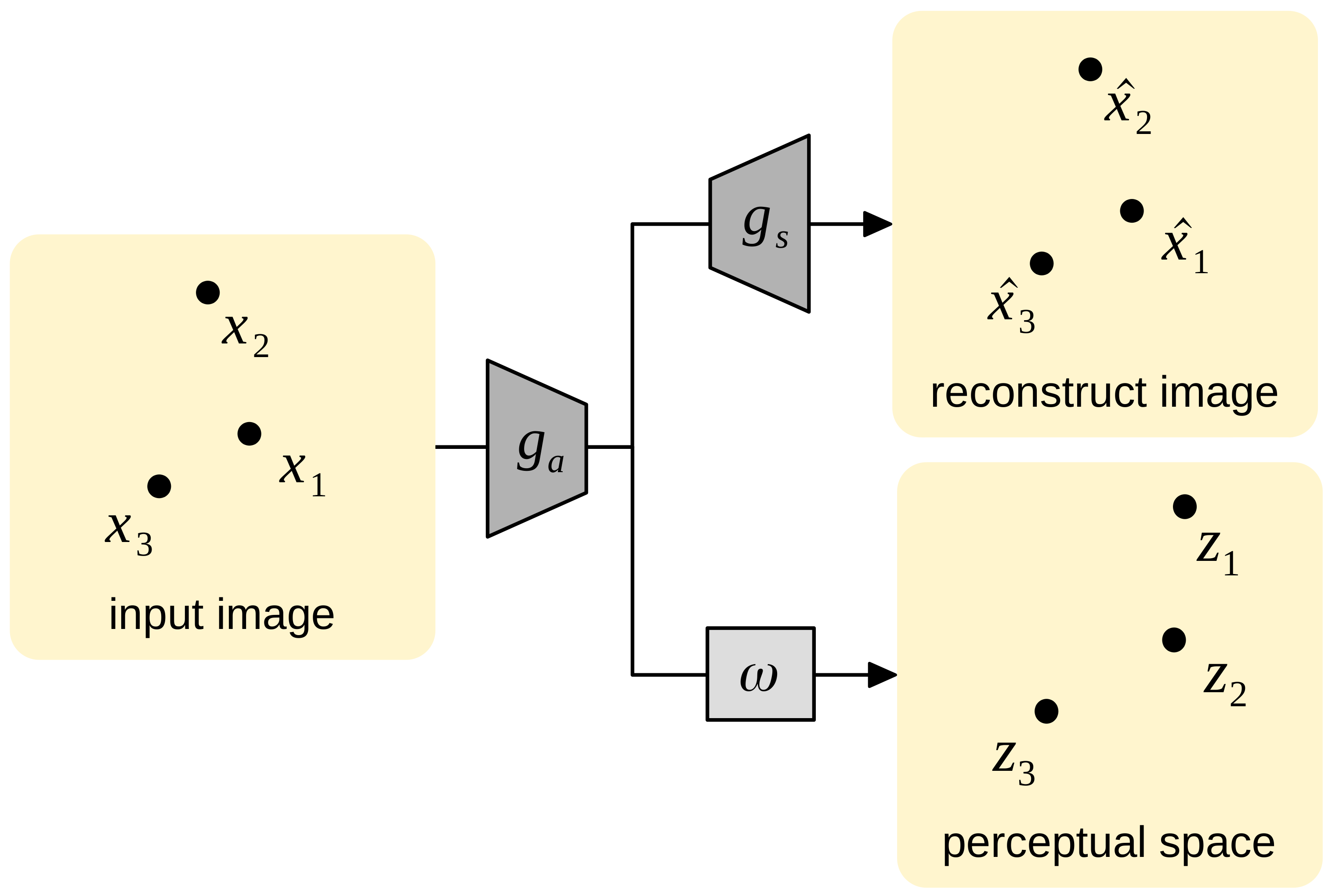} } \label{fig:flow2} }%
\caption{(a) The use of semantic and perceptual transforms for object classification and image similarity judgment. (b) The use of an analysis transform as a proxy for perceptual transform: $\rho_p=\omega(g_a(x);w)$.} %
\end{figure}

We consider a natural image $x$ as a point in the signal space that triggers stimuli in the brain's sensory cortex through the visual system. The effectiveness of a deep neural network (DNN) lies in its ability to learn a complex transformation $\rho_s$ that maps stimuli $x_i$ to points $s_i$ in the semantic space for object classification, as illustrated in Figure \ref{fig:flow1}. Equivalently, research in image quality assessment aims to learn a perceptual mapping $\rho_p$ such that the Euclidean distance between points $z_i$ in the perceptual space highly correlates with human perception.

In the field of neural science \cite{yamins2016using}, large capacity goal-driven hierarchical convolutional neural networks (HCNNs) like the VGG network \cite{simonyan2014very}, which are trained for image classification, have been demonstrated to effectively model neural stimuli and population responses in higher visual cortical areas. Consequently, the VGG network can map images into a feature vector space where distances between vectors closely align with the image's structures and semantics. Rather than employing per-pixel difference loss, studies have shown that utilizing perceptual loss from VGG leads to superior image quality in various image transformation tasks, such as style transfer \cite{gatys2015neural,johnson2016perceptual} and super-resolution \cite{ledig2017photo}.

Zhang et al. \cite{zhang2018perceptual} introduced the Learned Perceptual Image Patch Similarity (LPIPS) metric, which involves learning a set of linear weights $w$ applied to HCNN-extracted vectors to align with human perceptual judgments. LPIPS represents the pioneering attempt to employ a semantic transformation as a proxy for perceptual transformation. Essentially, they repurpose a semantic model as a perceptual model, where $\rho_p=\omega(\rho_s(x); w)$, with $\omega$ representing the learned linear mapping. Huang et al. \cite{huang2024cpips} proposed the Compressed Perceptual Image Patch Similarity (CPIPS) metric, which incorporates both an analysis transform $g_a$ and a synthesis transform $g_s$ within an end-to-end learned image compression codec \cite{balle2018variational} as a goal-driven HCNN for image reconstruction. In CPIPS, the analysis transform is pre-trained on ImageNet object classification to achieve a certain level of accuracy. Subsequently, it is jointly optimized for rate-distortion loss and classification loss. The CPIPS affirms that with a mapping function $\omega$ applied to the compressed latent representation, we can conduct human perceptual distance judgments with an accuracy comparable to LPIPS. This work extends CPIPS further and can be conceptually illustrated in Figure \ref{fig:flow2}. We study the following two key questions:

1) To what extent can the image encoder network be effectively utilized from both perceptual and semantic perspectives?

2) What outstanding tasks are for a neural codec to serve as a perceptual space transform?

\Section{Related Works} \label{_Related}

\SubSection{Learned Image Compression} \label{__lic}

The end-to-end learned image compression approaches have been proposed in the literature, starting with Ballé et al. \cite{balle2018variational} that surpassed traditional codecs like JPEG
and JPEG 2000
regarding PSNR and SSIM metrics. Minnen et al. \cite{minnen2018joint} further improved coding efficiency by employing a joint autoregressive and hierarchical prior model, surpassing the performance of the HEVC \cite{lainema2016hevc} codec. Cheng et al. \cite{cheng2020learned} developed techniques that achieved comparable performance to the latest coding standard VVC \cite{ohm2018versatile}. Several comprehensive surveys and introduction papers \cite{mishra2022deep,yang2022introduction} have summarized these advancements in end-to-end learned compression.


\SubSection{Image Quality Assessment} \label{__iqa}

The evaluation of image coding quality relies on full-reference image quality assessment metrics, which gauge the similarity between the reconstructed and the original images as human observers perceive. In addition to the traditional pixel-based PSNR, metrics such as SSIM \cite{wang2004image}, GMSD \cite{xue2013gradient}, and NLPD \cite{laparra2016perceptual} have been widely employed. Recently, new DNN-based methods like LPIPS \cite{zhang2018perceptual}, PIM \cite{bhardwaj2020unsupervised}, and DISTS \cite{ding2020image} have been proposed. These approaches have demonstrated superior predictive performance for subjective image quality. Their efficacy has been confirmed through validation against human judgments on benchmark datasets like BAPPS \cite{zhang2018perceptual} and CLIC2021 \cite{clic2021iqa}, which comprise a comprehensive collection of two-alternative force choice (2AFC) human judgments.

\SubSection{Semantic Features as Perceptual Loss} \label{__perloss}

The feature space of the VGG network has been leveraged by Gatys et al. \cite{gatys2015neural} to generate high-quality images for style transfer. Johnson et al. \cite{johnson2016perceptual} further introduced the \textit{feature reconstruction loss} and \textit{style reconstruction loss} computed up to the VGG network layer \texttt{relu3\_3} and \texttt{relu4\_3}, respectively. Features extracted from higher VGG layers encapsulate high-level semantics, including color, texture, and shape, collectively representing an image's style. Johnson's work encompasses the joint optimization of content and style loss, expanding beyond style transfer to encompass generic image transformation tasks.

In the domain of image super-resolution, Ledig et al. \cite{ledig2017photo} introduced SRGAN, which incorporates adversarial training and a perceptual loss function comprising an adversarial loss and a content loss. To capture low-level image details, they define the content loss based on the VGG network up to the \texttt{relu2\_2} layer. SRGAN has demonstrated remarkable results in achieving visually appealing 4X super-resolution and has garnered superior mean opinion scores (MOS) compared to prior methods. In addition to these two image tasks, other image reconstruction tasks \cite{liu2018x} have also used perceptual loss to enhance the visual quality.

\Section{Proposed Method} \label{_Method}

Transform coding plays a crucial role in conventional image compression as it separates the task process of decorrelating a source from its coding process \cite{balle2020nonlinear}. Popular transforms such as the Discrete Cosine Transform (DCT) and the Wavelet transform are linear and unitary. Consider $U$ as a unitary transform in $Z=UX$, then the energy in the transform domain equals the energy in the original domain: $\lVert Z \rVert^2 = \lVert X \rVert^2$. If $z_1=Ux_1$ and $z_2=Ux_2$, we have

\begin{equation}
\lVert z_1-z_2 \rVert^2 = \lVert x_1-x_2 \rVert^2
\end{equation}

A unitary transform preserves Euclidean distances from the signal space to the latent space. As a result, leveraging the latent representation in traditional image compression for high-level semantic computation is equivalent to processing in the signal domain. However, the development of complex non-linear transforms through deep neural networks in modern learned image compression \cite{balle2018variational,minnen2018joint} has made it possible to perform processing and analysis in the compressed domain, giving rise to the advocacy of Video Coding for Machines (VCM) \cite{evidence2020vcm}.

\SubSection{CPIPS: An Image Encoder Tuned with Semantics} \label{__cpips}

\begin{figure}[!ht]
\centering
\includegraphics[width=\columnwidth]{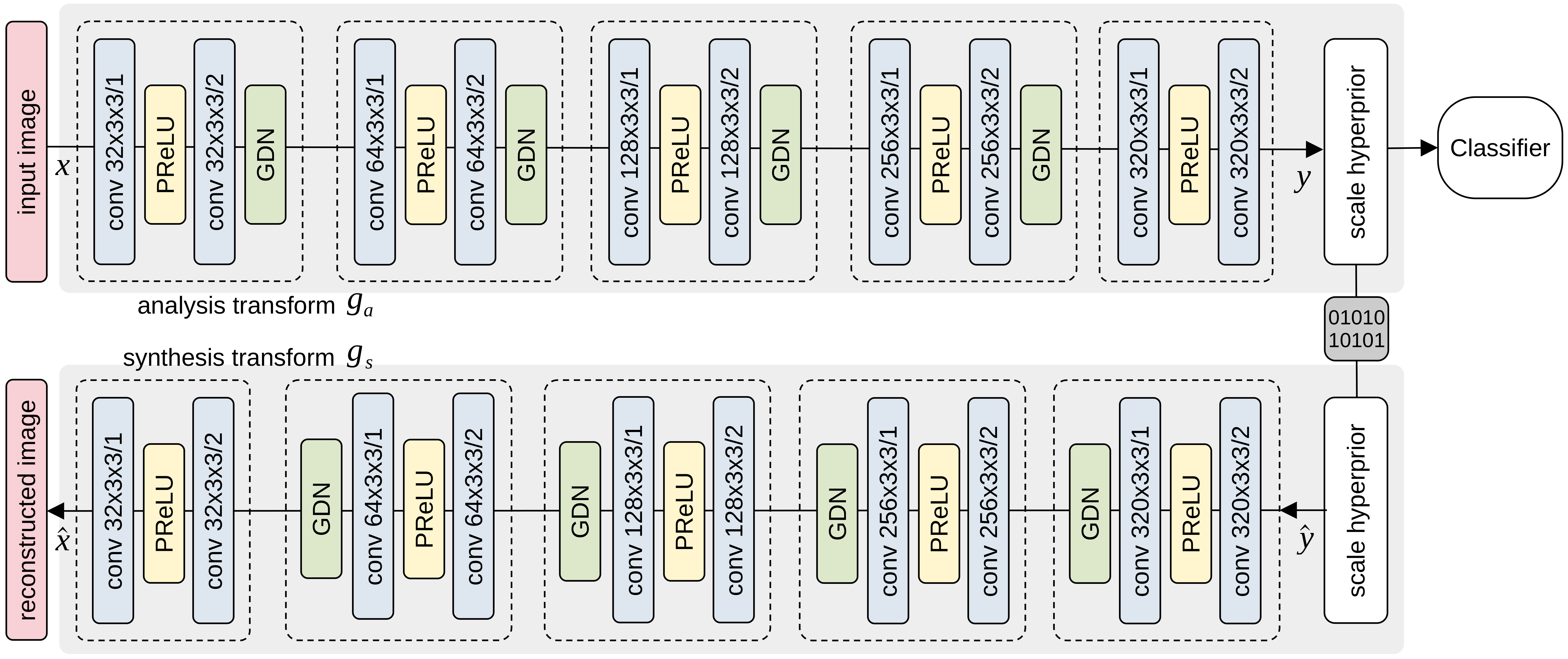}
\caption{The enhanced network architect of CPIPS. We use the parameterized ReLU and the Generalized Divisive Normalization (GDN) \cite{balle2018variational} as the activation function. The convolution notation $n\times k\times k/s$ represents the filter numbers, kernel size, and stride size.}
\label{fig:cpips-arch}
\end{figure}

In CPIPS \cite{huang2024cpips}, the weights of the analysis transform are initialized with the semantic features pre-trained on ImageNet. Through joint compression classification training, the gradient descent optimizer updates the encoder-decoder weights to analyze and synthesize the image while improving classification accuracy. We enhance CPIPS by removing the regularization loss and leveraging the encoder instead of the decoder. Figure \ref{fig:cpips-arch} shows the detailed network architecture.

To obtain the distance between two images, denoted as $x$ and $x_0$, we follow the same procedure as LPIPS \cite{zhang2018perceptual} by learning a linear layer $w$ on the BAPPS dataset. We extract feature maps $f^l \in \mathbb{R}^{C_l\times H_l\times W_l}$ for all layers $l$ and normalize them in the channel dimension. The activations are then scaled channel-wise using the vector $w_l \in \mathbb{R}^{C_l}$, and the $l_2$ distance is computed. Finally, we average across the spatial dimensions and sum up all layers for the final distance:

\begin{equation}
d(x,x_0) = \sum_l \frac{1}{H_lW_l} \sum_{h,w} \lVert w_l \odot (f_{hw}^l - f_{0hw}^l) \rVert_2^2 ,
\end{equation}

where $\odot$ denotes the convolution operator. We jointly train the compression classification loss to achieve 51.58\% and 76.27\% for top-1 and top-5 accuracy, respectively. The enhanced CPIPS performs superior to LPIPS on the BAPPS human perceptual judgment dataset, as shown in Table \ref{tab:perceptual}. However, its coding efficiency is somewhat compromised, surpassing only JPEG's.

\SubSection{Hyperprior encoder and Semantic-tuned Hyperprior} \label{__hyper}

A perceptual distance-preserving image encoder only makes sense when its coding efficiency remains superior to traditional hand-crafted codecs. To address this, we turn to the well-known hyperprior codec \cite{balle2018variational}, which has been reported to outperform HEVC in terms of PSNR metric. Given that image reconstruction is a well-defined goal-driven HCNN, we assess the feasibility of utilizing the hyperprior's analysis transform as the perceptual transform. Additionally, we apply the same joint compression-classification optimization strategy to fine-tune the hyperprior encoder/decoder as we wonder about the semantic-aware benefits derived from the auxiliary image classification task. Table \ref{tab:rd-gain} reports the Bjøntegaard-Delta (BD) metric compared to JPEG on the Kodak dataset. The fine-tuning of the hyperprior encoder significantly reduces its coding efficiency but remains superior to traditional codecs.

\begin{table}[!ht]
\centering
\caption{The coding efficiency impact of classification fine-tuning compared to JPEG.}
\resizebox{0.84\columnwidth}{!}{
\begin{tabular}{lrr|rr}
\hline
\textbf{Comparison} & \textbf{BD Rate (\%)} & \textbf{BD PSNR (dB)} & \textbf{Top-1 Acc.} & \textbf{Top-5 Acc.} \\
\hline
CPIPS & -1.17 & 0.54 & 51.58 & 76.27 \\
Hyperprior & -55.61 & 3.93 & \multicolumn{2}{c}{-} \\
Hyperprior-tune & -31.85 & 1.72 & 24.25 & 46.63 \\
\hline
\end{tabular} \label{tab:rd-gain}
}
\end{table}

\SubSection{Style Transfer and Super-resolution} \label{__img_trans}

In addition to perceptual similarity judgments, we assess the efficacy of the analysis transform as a perceptual loss for image transformation tasks. We employ Gatys' approach \cite{gatys2015neural} and SRGAN's technique \cite{ledig2017photo} for style transfer and super-resolution, utilizing CPIPS, hyperprior codec, and the hyperprior-tuned encoder as the perceptual loss networks. The feature layers we extract are detailed in Table \ref{tab:net-layer}. We focus on early layers from the analysis transform to capture low-level structures as content loss for style transfer. Conversely, we extract full layers to represent semantic features in super-resolution, as the original paper suggests. Each encoder network includes a final Gaussian-conditioned hyperprior bottleneck layer, reflecting the rate constraint imposed on the latent vectors.

\begin{table}[!ht]
\centering
\caption{The feature extraction layers for style transfer and super-resolution.}
\resizebox{0.6\columnwidth}{!}{
\begin{tabular}{lll}
\hline
\multirow{2}{*}{\textbf{Network}} & \multicolumn{2}{c}{\textbf{Feature Extraction Layer}} \\
 & \multicolumn{1}{c}{\textbf{Style Transfer}}  & \multicolumn{1}{c}{\textbf{SRGAN}}  \\
\hline
Vgg16 & \texttt{conv2\_2+ReLU} & \texttt{conv5\_3+ReLU}  \\
CPIPS & \texttt{conv2\_2+GDN} & \texttt{conv5\_2+bottleneck}  \\
Hyperprior & \texttt{conv2+GDN} & \texttt{conv4+bottleneck} \\
Hyperprior-tune & \texttt{conv2+GDN} & \texttt{conv4+bottleneck} \\
\hline
\end{tabular} \label{tab:net-layer}
}
\end{table}

\Section{Experimental Results} \label{_Expr}


For our proposed CPIPS, we begin by pre-training the analysis transform on the ImageNet dataset for 90 epochs, focusing on an image classification task, resulting in a top-1 accuracy of 59.68\%. Subsequently, we jointly train the compression classification task, utilizing the pre-trained weights for an additional 120 epochs. This phase involves employing the Adam optimizer with a learning rate set to 0.0001. In the case of the hyperprior method, we use the CompressAI implementation\footnote{https://github.com/InterDigitalInc/CompressAI} along with a pre-trained model configured at quality setting 8. We apply the same joint compression-classification optimization strategy to fine-tune the hyperprior codec, conducting this process for 120 epochs, resulting in a semantic-tuned hyperprior, denoted as Hyperprior-tune.

To evaluate our approach, we compare the perceptual judgment results with various quality metrics, namely GMSD, NLPD, LPIPS\footnote{https://github.com/richzhang/PerceptualSimilarity}, PIM\footnote{https://github.com/google-research/perceptual-quality}, and DISTS\footnote{https://github.com/dingkeyan93/DISTS}, on two datasets: BAPPS \cite{zhang2018perceptual} and CLIC2021 \cite{clic2021iqa}. These datasets consist of 151k and 122k patches, respectively, and employ a 2AFC judgment methodology.

For style transfer, we adopt a separate model for each style, fine-tuning the weights for content and style loss based on the activation magnitude of different networks to achieve visually appealing results. In the case of super-resolution, we conduct experiments on three widely recognized benchmark datasets: Set5, Set14, and BSD100. All experiments are performed with a scale factor of 4X between low and high-resolution images.

\SubSection{Perceptual Distance Judgments} \label{__perceptual}

The quality metrics are compared on the benchmark datasets, as shown in Table \ref{tab:perceptual}. It is evident that there exists a significant gap between modern DNN-based quality metrics and traditional pixel-based metrics. The PIM exhibits the best performance among specially designed perceptual metrics, and the DISTS metric also outperforms the LPIPS. Notably, the DISTS metric also leverages the VGG network as a fundamental building block.

\begin{table}[!ht]
\centering
\caption{The accuracy of human perceptual distance compared with various metrics. We boldly mark the numbers for leading group performance and underline the best.}
\resizebox{0.9\columnwidth}{!}{
\begin{tabular}{l|rrrrrrr|r}
\hline
\multirow{2}{*}{\textbf{Method}} & \multicolumn{7}{c|}{\textbf{BAPPS-2AFC}} & \multirow{2}{*}{\textbf{CLIC}} \\ 
 & \textbf{Trad.} & \textbf{CNN} & \textbf{S.Res} & \textbf{DeBlur} & \textbf{Color} &  \textbf{F.Interp} & \textbf{Avg.} & \\ \hline
LPIPS-Vgg \cite{zhang2018perceptual} & 73.36 & 82.20 & 69.51 & 59.43 & 61.62 & 62.34 & 68.08 & 61.06 \\
CPIPS \cite{huang2024cpips} & \textbf{77.08} & \textbf{83.33} & \textbf{71.18} & 60.25 & \textbf{63.63} & 61.62 & \textbf{69.52} & 60.92\\
CPIPS-hyper & 72.76 & 82.78 & 70.62 & \textbf{60.45} & 62.64 & \textbf{62.37} & 68.60 & \textbf{62.16} \\
CPIPS-hyper-tune & 72.98 & \textbf{82.79} & \textbf{71.50} & \textbf{60.29} & 62.00 & 62.27 & 68.64 & \textbf{61.73} \\
DISTS \cite{ding2020image} & \textbf{77.18} & 82.18 & 71.03 & 60.03 & \textbf{62.67} & \textbf{62.48} & \textbf{69.26} & 61.22 \\
PIM \cite{bhardwaj2020unsupervised} & \underline{\textbf{80.11}} & \underline{\textbf{87.37}}  & \underline{\textbf{76.01}} & \underline{\textbf{66.84}} & \underline{\textbf{70.53}} & \underline{\textbf{66.68}} & \underline{\textbf{74.59}} & \underline{\textbf{65.06}} \\ \hline
GMSD \cite{xue2013gradient} & 58.81 & 79.41 & 66.39 & 58.84 & 57.74 & 56.47 & 62.94 & 53.55 \\
NLPD \cite{laparra2016perceptual} & 56.81 & 79.09 & 64.85 & 58.26 & 56.03 & 55.41 & 61.74 & 53.05 \\
PSNR & 59.94 & 77.76 & 64.67 & 58.19 & 63.50 & 55.02 & 63.18 & 54.57\\
SSIM & 62.73 & 77.59 & 63.13 & 54.23 & 60.88 & 57.10 & 62.62 &55.05\\
\hline
\end{tabular} \label{tab:perceptual}
}
\end{table}

The off-the-shelf perceptual model, CPIPS, closely trails behind a custom-tailored quality metric like DISTS. Our CPIPS, built on a neural encoder, performs slightly better than DISTS in the BAPPS dataset but is less competitive in the CLIC dataset. Surprisingly, the CPIPS built on the original hyperprior codec, referred to as CPIPS-hyper, delivers a competitive performance in the BAPPS dataset and is superior in the CLIC dataset, closely rivaling a well-designed quality metric like DISTS. We argue that this phenomenon echoes the conclusion from \cite{yamins2016using}, where a well-trained neural network on an image reconstruction task is equivalent to a goal-driven HCNN. Another finding is that fine-tuning the joint compression classification optimization on the hyperprior encoder does not have a meaningful impact on perceptual prediction performance. We attribute this to the network architecture design problem: a pre-trained network good at classification like CPIPS is challenging to achieve satisfying coding efficiency while fine-tuning a highly efficient hyperprior codec cannot improve perceptual prediction. Investigating and improving upon the above findings are left as future work. Despite this, we confirmed that a vanilla hyperprior encoder performs closely, rivaling a well-designed quality metric like DISTS.

\SubSection{Style Transfer} \label{__style}

\begin{figure}[!ht]
\centering
\includegraphics[width=0.9\columnwidth]{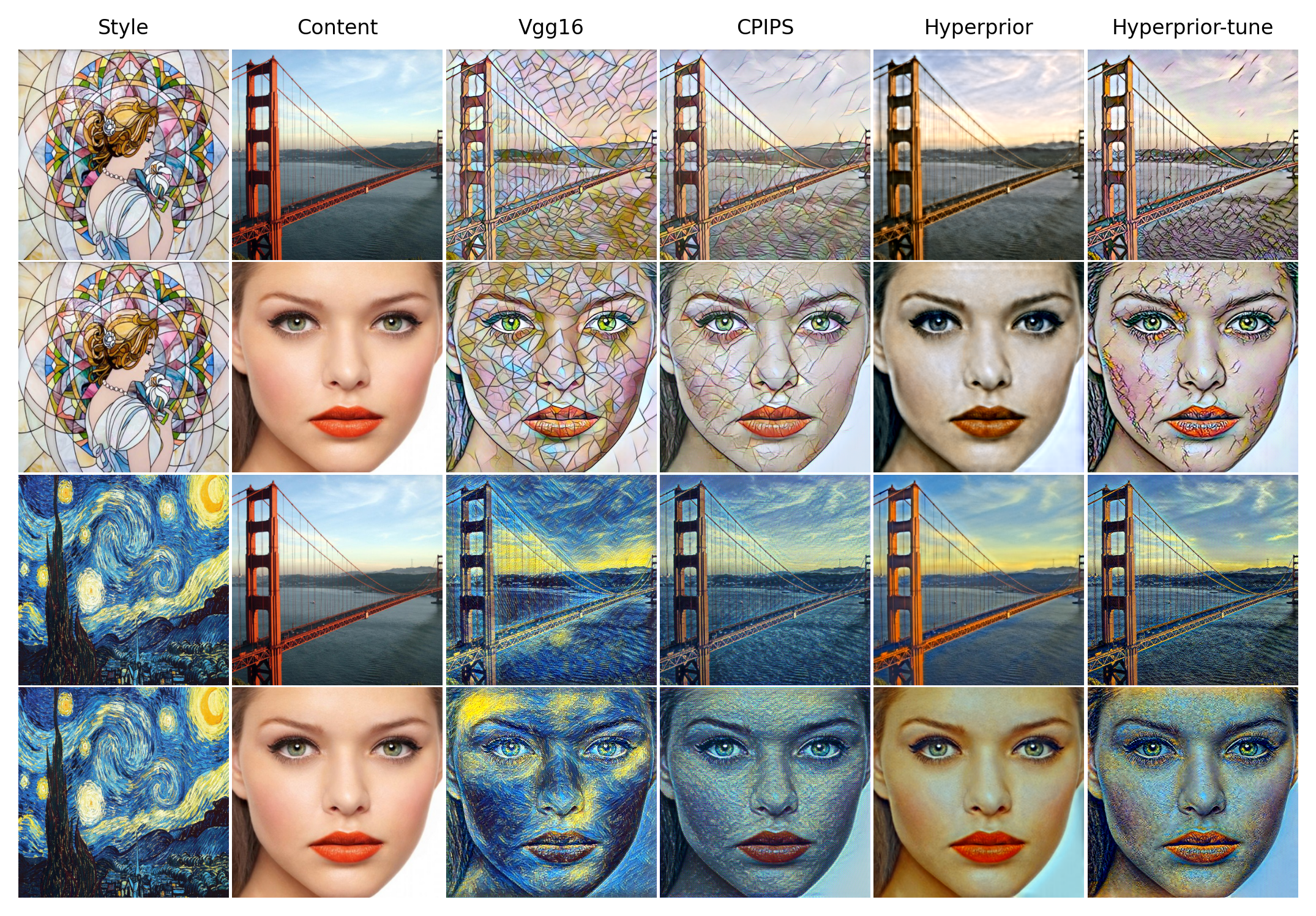}
\caption{Qualitative comparisons of style transfer. The transferred images' snapshots (started from the third column) show that the raw hyperprior codec struggles to transfer high-level features as styles.}
\label{fig:style-quality}
\end{figure}

We assess the analysis transform's ability to extract high-level features for style and transfer them to another content image. Figure \ref{fig:style-quality} showcases the qualitative results obtained using different networks as a perceptual loss. The CPIPS encoder effectively disentangles content and style features through semantic fine-tuning, allowing for style transfer similar to the VGG network. However, the stylish effect, while present, may not be as pronounced as with VGG, according to subjective evaluation. On the other hand, the raw hyperprior encoder struggles to differentiate between style and content features, resulting in less meaningful outputs. By applying semantic fine-tuning similar to CPIPS, the hyperprior's performance is greatly enhanced, producing stylish results comparable to those of CPIPS. Therefore, the vanilla neural encoder needs to be fine-tuned to accomplish the style transfer task.

\SubSection{Super-resolution} \label{__sr}

While it is acknowledged that SRGAN, with a perceptual loss, yields higher mean square errors, it often achieves superior mean opinion scores. We begin by quantifying and presenting the objective measurements in Table \ref{tab:sr-result}. When utilizing the CPIPS encoder as the perceptual loss, we obtain quantitative results comparable to those achieved with the VGG network. Consequently, the raw and tuned hyperprior encoder networks outperform the VGG network regarding PSNR and SSIM metrics. This result could be attributed to the fact that neural codecs are designed to excel in image reconstruction tasks.

\begin{table}[!ht]
\centering
\caption{Quantitative results for SRGAN 4X super-resolution.}
\resizebox{0.8\columnwidth}{!}{
\begin{tabular}{lrrrrrrr}
\hline
\multirow{2}{*}{\textbf{Method}} & \multicolumn{2}{c}{\textbf{Set5}} & \multicolumn{2}{c}{\textbf{Set14}} & \multicolumn{2}{c}{\textbf{BSD100}} \\ 
 & \textbf{PSNR} & \textbf{SSIM} & \textbf{PSNR} & \textbf{SSIM} & \textbf{PSNR} & \textbf{SSIM} & \\ \hline
Vgg16 & 28.87 & 0.8371 & 25.82 & 0.7256 & 25.78 & 0.7016\\
CPIPS \cite{huang2024cpips} & 28.77 & 0.8374 & 25.73 & 0.7241 & 25.75 & 0.7024\\
CPIPS-hyper & \textbf{29.30} & \textbf{0.8537} & \textbf{26.06} & \textbf{0.7396} & \textbf{25.98} & \textbf{0.7128} \\
CPIPS-hyper-tune & \textbf{29.27} & \textbf{0.8534} & \textbf{26.04} & \textbf{0.7389} & \textbf{25.97} & \textbf{0.7118} \\
\hline
\end{tabular} \label{tab:sr-result}
}
\end{table}

\begin{figure}[!ht]
\centering
\includegraphics[width=0.9\columnwidth]{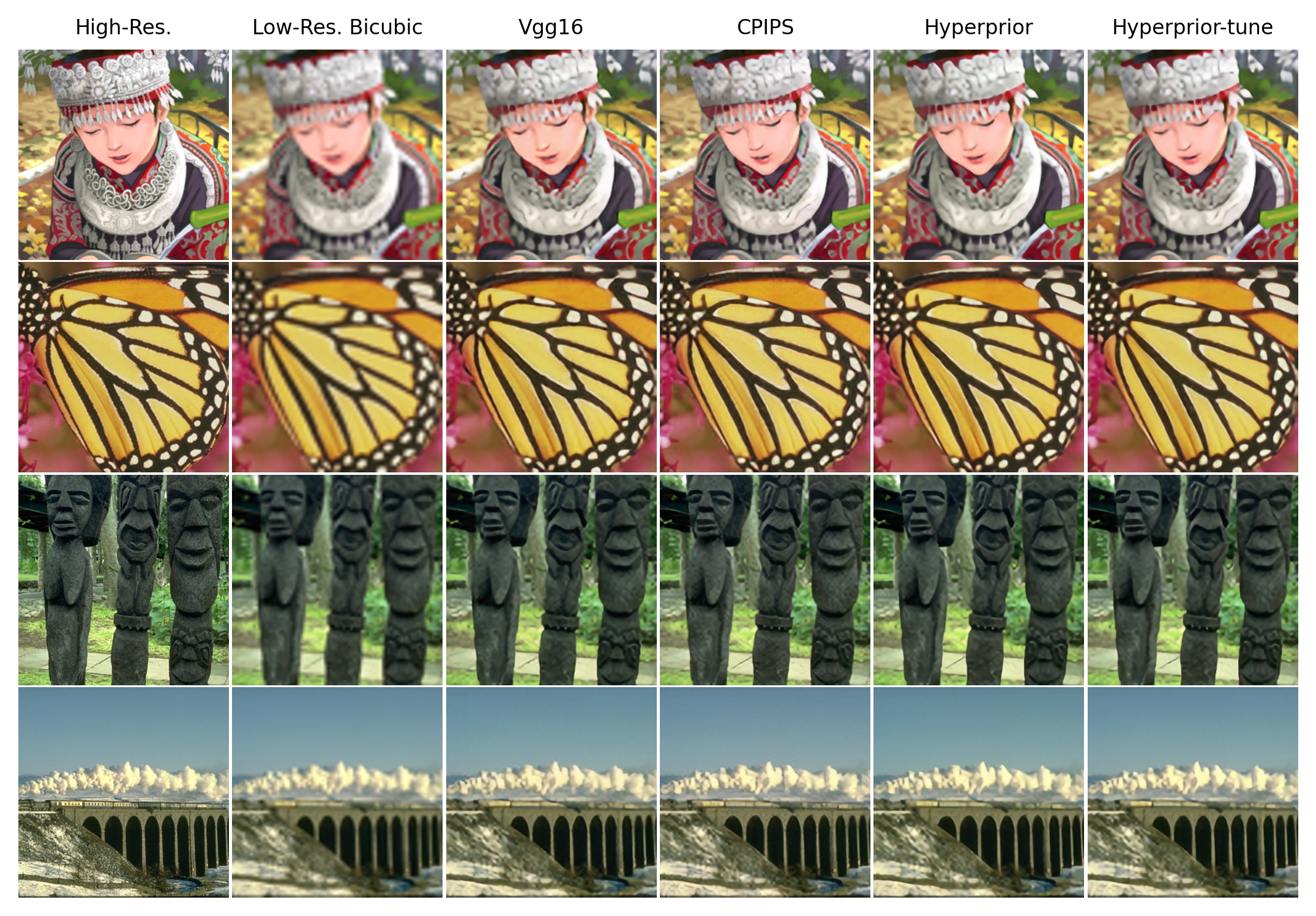}
\caption{Visual comparisons of SRGAN 4X super-resolution.}
\label{fig:sr-quality}
\end{figure}

In terms of visual quality, as illustrated in Figure \ref{fig:sr-quality}, all the analysis transforms yield sharpened and visually enhanced outputs. Remarkably, using the VGG network introduces finer details than other methods. This observation suggests that the analysis transform derived from a neural codec could be a reliable proxy for perceptual loss in specific image transformation tasks.

\SubSection{Rate-Distortion Trade-offs} \label{__rate_distortion}

In lossy compression, defining distinct quality factors for low and high-bit-rate applications through rate-distortion optimization is a common practice. We evaluated the influence of bit-rate trade-offs on the efficacy of perceptual representation. Our experiments revealed that even with the lowest bit-rate setting, such as quality 1, the accuracy of perceptual judgment, style transfer effectiveness, and super-resolution quality remained remarkably close to that achieved with a high-quality setting.



\Section{Conclusion} \label{_Conclude}

This study demonstrated that the analysis transform within a neural codec is highly proficient in extracting feature representations suitable for perceptual distance judgment and various image transformation tasks. Specifically, we've shown its effectiveness as a perceptual loss in super-resolution tasks, generating visually pleasing outputs with or without classification fine-tuning. In the case of style transfer, fine-tuning the analysis transform is crucial for producing stylish content. Suppose one seeks to repurpose the neural encoder as a perceptual transform. In that case, it's imperative to carefully re-examine the network architecture design, pre-training techniques, and rate-distortion-semantic optimization. Further exploration into auxiliary tasks beyond image classification may also serve as avenues for semantically fine-tuning the neural encoder.

\Section{References}
\bibliographystyle{IEEEbib}
\bibliography{all_refs}

\end{document}